%% file: main.tex
\documentclass[runningheads,a4paper]{llncs}

\usepackage[british]{babel}

\usepackage{hyperref}

\usepackage{amssymb}
\usepackage{amsmath}

\usepackage{wrapfig}

\usepackage{mathptmx}
\usepackage{graphicx}
\usepackage{pdflscape}
\makeatletter
\providecommand{\bigsqcap}{%
\mathop{%
\mathpalette\@updown\bigsqcup
}%
}
\newcommand*{\@updown}[2]{%
  \rotatebox[origin=c]{180}{$\m@th#1#2$}%
}
\makeatother

\usepackage{tikz}

\begin{document}
\mainmatter 

\title{Scene Learning, Recognition and Similarity Detection in a Fuzzy Ontology via Human Examples}
\titlerunning{Scene Learning, Recognition and Similarity Detection} 
\toctitle{Scene Learning, Recognition and Similarity Detection in a Fuzzy Ontology via Human Examples}

\author{Luca Buoncompagni\inst{1} \and Fulvio Mastrogiovanni\inst{1} \and Alessandro Saffiotti\inst{2}}
\authorrunning{Luca Buoncompagni, Fulvio Mastrogiovanni and Alessandro Saffiotti}
\tocauthor{Luca Buoncompagni, Fulvio Mastrogiovanni and Alessandro Saffiotti}

\institute{
Department of Informatics, Bioengineering, Robotics and Systems Engineering,\\University of Genoa, Italy\\ 
\{\href{mailto:luca.buoncompagni@edu.unige.it}{luca.buoncompagni@edu.}~,~\href{mailto:fulvio.mastrogiovanni@unige.it}{fulvio.mastrogiovanni@}\}unige.it
\and
Centre for Applied Autonomous Sensor Systems, \"{O}rebro University, Sweden\\ 
\href{mailto:asaffio@aass.oru.se}{asaffio@aass.oru.se}
}
% typeset the title of the contribution
\maketitle

\begin{abstract}
%% (fuzzySIT\repository can...) (fuzzydl, type of perception, simulation)
This paper introduces a Fuzzy Logic framework for scene learning, recognition and similarity detection, where scenes are taught via human examples.
The framework allows a robot to:
(i) deal with the intrinsic \textit{vagueness} associated with determining spatial relations among objects;
(ii) infer similarities and dissimilarities in a set of scenes, and represent them in a hierarchical structure represented in a Fuzzy ontology.
In this paper, we briefly formalize our approach and we provide a few use cases by way of illustration. Nevertheless, we discuss how the framework can be used in real-world scenarios.
\end{abstract}

\section{Introduction}
\label{sec:introduction}

%% purpose: (generalized recognition from human demonstrations, planning and memory retrieval)
In order to achieve a natural interaction between humans and robots, it is crucial robots be able not only to learn by human examples, but also to organize learned knowledge for a long-term interaction, as well as for communicating it to humans. 
On the one hand, such learning techniques as Learning by Example \cite{PbD}\cite{Duanetal2017} are efficient ways to teach robots well-defined skills, but the resulting representation typically does not take interaction aspects into account.
On the other hand, as any form of inductive reasoning, learning can deal with generalisation and the conceptualisation of knowledge only to a limited extent \cite{ML}, and therefore methods to support robot-to-human communication and vagueness in scene descriptions must be considered.

%% application: (table-top, map-representation)
In this paper, we present a perception framework\footnote{\label{note:repo}A first implementation is available at: \url{https://github.com/EmaroLab/fuzzy_sit}.} structurally requiring human examples and representing the environment using a formalism based on Fuzzy Logic. 
The approach allows a robot to acquire scenes of the robot's workspace via human examples, represent spatial relations among objects therein using fuzzy concepts, and hierarchically classify scenes on the basis of their similarities.  
We focus on scenes related to a tabletop scenario, which are sequentially shown by a human teacher. 
Our approach populates a fuzzy \emph{ontology}, which can be used to encode vague representations of spatial relations and reason upon them to perform classification, ground human-robot communication and, in perspective, perform task planning.  

%% objective: (of the paper, brief verbal TOC)
In the following Sections, we briefly introduce our framework and we show a few simple examples to describe the system's behaviour and discuss its main features.

\section{Method}
\label{sec:method}

%% framework components and input
Our framework is based on a fuzzy OWL ontology \cite{OWL}, managed by the fuzzyDL reasoner \cite{fuzzydl}, which (i) represents scenes in terms of fuzzy objects and fuzzy spatial relations among them, and (ii) determines the similarity between any pair of scenes using their fuzzy descriptions.

%In details, we develop two mapping procedures to manipulate such a representation at run-time and augment the knowledge by interactions.

We assume as an input (i) the classification of perceived objects according to a number of predefined fuzzy sets (e.g., \texttt{Book} or \texttt{Cup}), i.e., the object \textit{type}, and (ii) the knowledge of the spatial relations objects are involved in (e.g., \texttt{right} or \texttt{left}), both with the associated degree of membership, as shown in Table \ref{table:exp}.
Two \textit{mapping} procedures are used to manipulate the ontology at runtime:
$\mathbb{M}_1$ maps percepts to a fuzzy scene \textit{individual} \texttt{S} in the ontology, whereas
$\mathbb{M}_2$ creates a new fuzzy scene \textit{class} \texttt{Scene} from a single individual \texttt{S} would the latter not be classified by any class in the ontology.

In order to understand how $\mathbb{M}_1$ works, let us assume that the robot perceives a scene where two books, namely \texttt{B} and \texttt{D}, are detected, with \texttt{D} at the right hand side of \texttt{B} (first scenario in Section \ref{table:exp}).
A fuzzy set \texttt{Book} is defined to assess the degree of membership of the two books \texttt{B} and \texttt{D}, for instance \texttt{Book(B,1)} and \texttt{Book(D,.8)}.
A fuzzy spatial relation \texttt{right} is used to represent the degree of membership of the assertion \texttt{D} \textit{at the right hand side of} \texttt{B}, for instance \texttt{right(B,D,.9)}.
Then, for all objects which are in a \texttt{right} relationship with other objects, we sum up the degrees of membership of the corresponding fuzzy relation, in this case \texttt{.9}, to determine a reified description relating object types and relations \cite{reify}, e.g., \texttt{hasBookRight(.8)}, which we use to define \texttt{S}. 
In general terms:
\begin{equation}
\label{eq:S}
\texttt{S} \triangleq \bigsqcap_{\substack{j \in \bar{\Lambda},\,k \in \bar{\lambda}}} \texttt{has}\Delta_{jk}.e_{jk},
\end{equation}
where:
\begin{equation}
\label{eq:S_specs}
e_{jk} = \sum \Lambda_j \otimes \exists \lambda_k.\Lambda_j
\equiv \sum_{\substack{Y \in \bar{\Lambda}}} \max_{I \in \Lambda_j} \left\{ \min \left\{ \bar{\Lambda}(I), Y.\lambda_k(I) \right\} \right\}.
\end{equation}
In \eqref{eq:S} and \eqref{eq:S_specs},  $\bar{\Lambda}=\{\Lambda_1,\ldots,\Lambda_j,\ldots, \Lambda_N\}$ is the set of object types represented as fuzzy sets (e.g., \texttt{Book} and \texttt{Cup}), $\bar{\lambda} = \{\lambda_1,\ldots,\lambda_k,\ldots,\lambda_M\}$ is the set of spatial relations represented as fuzzy relations (e.g., \texttt{right}), and $\Delta_{jk}$ is the reification of ${\Lambda_j}$ on ${\lambda_k}$, e.g., \texttt{BookRight}.

%% scene relation count: (sigma count and restriction)
We assess the similarity between two scenes by comparing the number $e_{jk}$ of relations involving a certain object type (e.g., \texttt{Book} and \texttt{right}) in scene descriptions in the form of \eqref{eq:S}.
If we considered a non fuzzy formulation, this would mean defining a set of minimal cardinality restrictions over the definition of \texttt{S} (e.g., \emph{the scene has at least one book on the right hand side}).
However, this is not the case with a fuzzy formulation \cite{fuzzyRestriction}.
In our case, we adopt the Sigma Counter approach \cite{sigmaCounter}, and we compare the sum of all degrees of membership for a given fuzzy spatial relation with respect a given object type with a \textit{left-shoulder} membership function, which restricts the counter value through classification, i.e., the apex of the shoulder for which its value becomes $1$ occurs at $e_{jk}$.

%% scene class: (mapping, dimensions and recognition)
$\mathbb{M}_2$ determines Sigma Counter restrictions for all the reified spatial relations in the set $\bar{\Delta}$. 
When all restrictions are computed, a fuzzy scene class \texttt{Scene} is created as a fuzzy set whose purpose is to classify \texttt{S}.
Considering the example introduced above, a description is created such that \texttt{hasBookRight.atLeast(.8)}, and it is assigned to \texttt{Scene}.
In formulas:
\begin{equation}
\label{eq:learn}
\texttt{Scene} \equiv \bigotimes_{\substack{j \in \bar{\Lambda},\,k \in \bar{\lambda}}} \exists \texttt{has}\Delta_{jk}.\texttt{atLeast}(LeftShoulder(e_{jk})).
\end{equation}

Three remarks can be made:
(i) a class is inductively derived from a single example, when the fuzzy ontology does not contain any fuzzy set whose description is suitable to classify a scene individual; 
(ii) human supervision may be needed to corroborate or modify the class description;
(iii) the scene formulation in \eqref{eq:learn} allows for reasoning about similarities among scenes; 
the fuzzyDL reasoner proves able to build a hierarchy of scene individuals using the fuzzy matching of all the minimal cardinality restrictions, which acts as a sort of fuzzy implication between scenes;
an example is shown in \autoref{fig:hierarchy}, and discussed below.

\section{Preliminary Results and Discussion}
\label{sec:results}

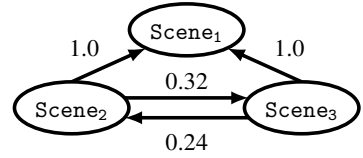
\begin{wrapfigure}{R}{0.4\textwidth}
\centering
  
\scalebox{.9}{\input{hierarchy.tex}}

\vspace{-2.5em}
\caption{The scene hierarchy obtained as a result of the learning process: circles are fuzzy scene classes, while arrows are fuzzy implications with the related degree of membership.}
\label{fig:hierarchy}
\end{wrapfigure}

Experiments have been performed in an incremental manner.
At bootstrap, the ontology does not contain assertions. 
Then, all scenes are shown sequentially to the robot, which applies the mappings $\mathbb{M}_1$ and $\mathbb{M}_2$ described above. 
We assume a fuzzy scene individual as being recognised when the associated degree of membership is higher than $.97$.
When all scenes are classified, a hierarchy is induced, which is shown in \autoref{fig:hierarchy}.
If we focus on the two last columns of Table \ref{table:exp}, it is possible to see how both learning and recognition performance scale with the complexity of descriptions, as well as with the number of assertions in the ontology.
All experiments have been performed using a graphical user interface (available open source\textsuperscript{\ref{note:repo}}), which simulates the perception of objects and spatial relations.
Results have been collected using an Intel Core i5-460M ($2.53$ $Ghz$) processor with $4$ $GB$ of $DDR3$ memory.

%% table explanation
Table \ref{table:exp} shows a few examples of scenes presented to the robot, where only two object types and one spatial relation is considered.
The first column of the table depicts object arrangements, whereas the second and the third column represent the fuzzy \textit{predicates} specifying the degrees of memberships for object types and spatial relations.
The fourth column shows the description of the resulting fuzzy scene individual, obtained applying \eqref{eq:S} and \eqref{eq:S_specs}, whereas in the fifth column the corresponding description of a fuzzy scene class is shown. 
The last two columns provide an indicative estimate of learning and recognition times, computed respectively when a fuzzy scene class is first created and when another individual is subsequently classified as being an instance of that class.

%% experiment table
\input{table.tex}

%% increment learning and performances
In the recognition phase, a new fuzzy scene individual \texttt{S} can be classified as being an instance of one of three fuzzy scene classes (i.e., $\texttt{Scene}_\texttt{1}$, $\texttt{Scene}_\texttt{2}$ and $\texttt{Scene}_\texttt{3}$), whereas in the  learning phase the hierarchy must be updated to create a new fuzzy scene class, which increases the overall computational time.
It is noteworthy that learning and classification performance depend also on the number of objects in the scene and the related spatial relations. 
     
%% hierarchy figure
\setlength{\intextsep}{5pt}%
\setlength{\columnsep}{5pt}%
     
%% example hierarchy and GUI
It is possible to observe that if $\texttt{Scene}_\texttt{2}$ or $\texttt{Scene}_\texttt{3}$ are classified, also the objects arrangement in $\texttt{Scene}_\texttt{1}$ holds with a full degree, given the qualitative nature of spatial relations.
Furthermore, $\texttt{Scene}_\texttt{2}$ and $\texttt{Scene}_\texttt{3}$ share some similarity, even if not as clearly as in the previous case, which is represented by the reasoner with a low fuzzy implication value between them.
Such a behavior is due to the Sigma Count approach, specifically to how the left-shoulder function changes between 0 and $e_{jk}$.

%% example of recognition with vague perception
Table \ref{table:exp} shows also how the fuzzyDL reasoner can deal with vague spatial relations.
It is noteworthy that, just for the sake of argument, degrees of membership in spatial relations are negatively correlated with the actual distance between objects.
Finally, the framework can also deal with inaccurate scene recognitions.
It could be the case that an object $\texttt{B}$ is classified as a $\texttt{Book}(\texttt{B},.8)$ \textit{and also} as a $\texttt{Cup}(\texttt{B},.2)$, which is \textit{slightly} located on the left hand side of $\texttt{D}$, i.e., \texttt{right(D,B,.3)}.
In this case, applying \eqref{eq:S} gives:
$e_{\texttt{Book},\texttt{right}}\leqslant\max\left\{\min\left\{.2,.3\right\},\min\left\{.8,.3\right\}\right\}$.

\section{Conclusions and Future Work}
\label{sec:conclusions}

We introduce a framework based on Fuzzy Logic to learn, recognise and hierarchically classify tabletop scenes presented to the robot by a human teacher.
The framework can serve as a basis to ground human-robot communication processes: 
on the one hand, it can deal with the intrinsic vagueness associated with human perception of spatial relationships; 
on the other hand, its performance is sufficiently good for a natural interaction in human-robot interaction scenarios. 

%% further work: (performances + recreation)
The framework is under test to benchmark its scalability and representation capabilities.
Currently, research activities focus on:
(i) determining possible \textit{singularities} in the scene representation to avoid degenerate cases;
(ii) integration between \textit{natural} spatial representations and speech-based interaction;
(iii) evaluation of action planning methods to perform object manipulation with the aim of recreating a previously perceived scene.

\section*{Acknowledgements}
\label{sec:ack}
This work has been partly supported by a grant of the Fondazione/Stiftelsen C.M. Lerici awarded to the first author.

%\clearpage
\bibliography{bib}

\end{document}

%% file: table.tex
\begin{landscape}
\begin{table}[t!]
\caption{Three experiments sequentially performed within a simplified scenario.}
\centering
     \centerline{
          \begin{tabular}{c|c|c|c|c|cc}
               \multicolumn{3}{c|}{\textbf{Inputs}}         & 
               \multicolumn{2}{c|}{\textbf{Maps}} & 
               \multicolumn{2}{c} {\textbf{Performance}}     \\
     %%%%%%%%%%%%%%%%%%%%%%%%%%%%%%%%%%%%%%%%%%%%%%%%%%%%%%%%%%%%%%%%%%%%%%%%%%
               \textit{Scenario}                      & 
               \textit{Object Type}                   & 
               \textit{Spatial Relation}              & 
               \textit{Scene individual}              & 
               \textit{Scene class}                   & 
               \multicolumn{1}{c|}{\textit{Learning [$ms$]}} & 
               \textit{Recognition [$ms$]}         \\ \hline
     %%%%%%%%%%%%%%%%%%%%%%%%%%%%%%%%%%%%%%%%%%%%%%%%%%%%%%%%%%%%%%%%%%%%%%%%%%
     %%%%%%%%%%%%%%%%%%%%%%%%%%%%%%%%%%%%%%%%%%%%%%%%%%%%%%%%%%%%%%%%%%%%%%%%%%          
               \begin{tabular}[c]{@{}c@{}}
                      
\scalebox{1.2}{\input{s1.tex}}

               \end{tabular}               & 
               \begin{tabular}[c]{@{}c@{}}
                    \texttt{Book}(\texttt{B},\texttt{1})\\ 
                    \texttt{Book}(\texttt{D},\texttt{.8})
               \end{tabular}               & 
               \begin{tabular}[c]{@{}c@{}}
                    \texttt{right}(\texttt{B},\texttt{D},\texttt{.9})
               \end{tabular}               & 
               \begin{tabular}[c]{@{}c@{}}
                    \texttt{S} $\triangleq$ \\
                    \texttt{hasBookRight}(\texttt{.8})
               \end{tabular}               & 
               \begin{tabular}[c]{@{}c@{}}
                    $\texttt{Scene}_\texttt{1}$ $\equiv$ \\
                    \texttt{hasBookRight}.\texttt{atLeast}(\texttt{.8})
               \end{tabular}               & 
               \multicolumn{1}{c|}{1128} & 
               873                        \\ \hline
     %%%%%%%%%%%%%%%%%%%%%%%%%%%%%%%%%%%%%%%%%%%%%%%%%%%%%%%%%%%%%%%%%%%%%%%%%%          
               \begin{tabular}[c]{@{}c@{}}
                      
\scalebox{1.2}{\input{s2.tex}}

               \end{tabular}               & 
               \begin{tabular}[c]{@{}c@{}}
                    \texttt{Book}(\texttt{B},\texttt{.7})\\ 
                    \texttt{Book}(\texttt{D},\texttt{1})\\
                    \texttt{Cup}(\texttt{C},\texttt{.6})
               \end{tabular}               & 
               \begin{tabular}[c]{@{}c@{}}
                    \texttt{right}(\texttt{B},\texttt{C},\texttt{.5})\\
                    \texttt{right}(\texttt{C},\texttt{D},\texttt{.9})\\
                    \texttt{right}(\texttt{B},\texttt{D},\texttt{.3})
               \end{tabular}               & 
               \begin{tabular}[c]{@{}c@{}}
                    \texttt{S} $\triangleq$ \\
                    \texttt{hasBookRight}(\texttt{.9}+\texttt{.3})\\
                    $\sqcap$ \texttt{hasCupRight}(\texttt{.5})
               \end{tabular}               & 
               \begin{tabular}[c]{@{}c@{}}
                    $\texttt{Scene}_\texttt{2}$ $\equiv$ \\
                    \texttt{hasBookRight}.\texttt{atLeast}(\texttt{1.2})\\
                    $\otimes$ \texttt{hasCupRight}.\texttt{atLeast}(\texttt{.5})
               \end{tabular}               & 
               \multicolumn{1}{c|}{2381} & 
               2115                        \\ \hline     
     %%%%%%%%%%%%%%%%%%%%%%%%%%%%%%%%%%%%%%%%%%%%%%%%%%%%%%%%%%%%%%%%%%%%%%%%%%          
               \begin{tabular}[c]{@{}c@{}}
                      
\scalebox{1.2}{\input{s3.tex}}

               \end{tabular}               & 
               \begin{tabular}[c]{@{}c@{}}
                    \texttt{Book}(\texttt{B},\texttt{.8})\\ 
                    \texttt{Book}(\texttt{D},\texttt{1})\\
                    \texttt{Cup}(\texttt{C},\texttt{.7})
               \end{tabular}               & 
               \begin{tabular}[c]{@{}c@{}}
                    \texttt{right}(\texttt{B},\texttt{D},\texttt{1})\\
                    \texttt{right}(\texttt{B},\texttt{C},\texttt{.2})\\
                    \texttt{right}(\texttt{D},\texttt{C},\texttt{.5})
               \end{tabular}               & 
               \begin{tabular}[c]{@{}c@{}}
                    \texttt{S} $\triangleq$ \\
                    \texttt{hasBookRight}(\texttt{.8})\\
                    $\sqcap$ \texttt{hasCupRight}(\texttt{.2} + \texttt{.5})
               \end{tabular}               & 
               \begin{tabular}[c]{@{}c@{}}
                    $\texttt{Scene}_\texttt{3}$ $\equiv$ \\
                    \texttt{hasBookRight}.\texttt{atLeast}(\texttt{.8})\\
                    $\otimes$ \texttt{hasCupRight}.\texttt{atLeast}(\texttt{.7})
               \end{tabular}               & 
               \multicolumn{1}{c|}{3686} & 
               3389                        %\\ \hline 
     %%%%%%%%%%%%%%%%%%%%%%%%%%%%%%%%%%%%%%%%%%%%%%%%%%%%%%%%%%%%%%%%%%%%%%%%%%                             
%           \begin{tabular}[c]{@{}c@{}}
%                     \inputTikZ{1.2}{./image/s4.tex}
%                \end{tabular}               & 
%                \begin{tabular}[c]{@{}c@{}}
%                     ...                 
%                \end{tabular}               & 
%                \begin{tabular}[c]{@{}c@{}}
%                     ...
%                \end{tabular}               & 
%                \begin{tabular}[c]{@{}c@{}}
%                     ...
%                \end{tabular}               & 
%                \begin{tabular}[c]{@{}c@{}}
%                     ...
%                \end{tabular}               & 
%                \multicolumn{1}{c|}{...} & 
%                ...                        %\\ \hline               
          \end{tabular}
     }
\label{table:exp}
\end{table}
\end{landscape}

%% file: s1.tex
\definecolor{cffffff}{RGB}{255,255,255}

\begin{tikzpicture}[y=0.80pt, x=0.80pt, yscale=-1.000000, xscale=1.000000, inner sep=0pt, outer sep=0pt]
  \begin{scope}[cm={{0.20872,0.0,0.0,0.20872,(-4.15391,74.04193)}}]
  \end{scope}
  \path[xscale=0.998,yscale=1.002,fill=black,line join=miter,line cap=butt,line
    width=0.800pt] (19.5888,46.3275) node[above right] (text7140-9-2-0) {\texttt{B}};
  \path[xscale=0.998,yscale=1.002,fill=black,line join=miter,line cap=butt,line
    width=0.800pt] (58.0972,46.3275) node[above right] (text7140-9-2-0-73) {\texttt{D}};
  \begin{scope}[cm={{0.2703,0.0,0.0,0.2703,(9.40853,6.72892)}}]
    \path[color=black,draw=black,fill=black,line join=miter,line cap=butt,miter
      limit=4.00,even odd rule,line width=0.069pt] (42.8145,21.1191) --
      (42.8145,107.4219) -- (43.4648,107.4219) -- (61.8691,107.4219) --
      (61.8691,21.1191) -- (42.8145,21.1191) -- cycle(44.1152,22.4199) --
      (60.5684,22.4199) -- (60.5684,106.1231) -- (44.1152,106.1231) --
      (44.1152,22.4199) -- cycle;
    \begin{scope}[shift={(-0.20058,0)}]
      \path[draw=black,line join=miter,line cap=butt,miter limit=4.00,even odd
        rule,line width=0.520pt] (46.5085,29.5593) -- (58.5763,29.5593);
      \path[draw=black,line join=miter,line cap=butt,miter limit=4.00,even odd
        rule,line width=0.520pt] (46.5085,33.9720) -- (58.5763,33.9720);
      \path[draw=black,line join=miter,line cap=butt,miter limit=4.00,even odd
        rule,line width=0.520pt] (46.5085,38.3849) -- (58.5763,38.3849);
      \path[draw=black,line join=miter,line cap=butt,miter limit=4.00,even odd
        rule,line width=0.520pt] (46.5085,42.7977) -- (58.5763,42.7977);
    \end{scope}
    \path[fill=black,miter limit=4.00,even odd rule,line width=1.040pt]
      (52.3418,96.7458) circle (0.1588cm);
    \path[cm={{4.33325,0.0,0.0,4.33325,(-941.33439,-425.19092)}},draw=black,fill=cffffff,miter
      limit=4.00,even odd rule,line width=0.015pt] (227.4397,112.9677) --
      (227.4397,103.3397) -- (229.3021,103.3397) -- (231.1646,103.3397) --
      (231.1646,112.9677) -- (231.1646,122.5958) -- (229.3021,122.5958) --
      (227.4397,122.5958) -- (227.4397,112.9677) -- cycle(229.8108,121.6461) ..
      controls (229.8822,121.6197) and (230.0326,121.5206) .. (230.1450,121.4261) ..
      controls (230.9483,120.7501) and (230.6154,119.4131) .. (229.5890,119.1930) ..
      controls (228.6546,118.9926) and (227.8195,119.8427) .. (228.0616,120.7479) ..
      controls (228.2732,121.5391) and (229.0386,121.9321) .. (229.8108,121.6461) --
      cycle(230.7227,108.0117) .. controls (230.7227,107.9715) and
      (230.2176,107.9486) .. (229.3337,107.9486) .. controls (228.4498,107.9486) and
      (227.9447,107.9716) .. (227.9447,108.0117) .. controls (227.9447,108.0519) and
      (228.4498,108.0748) .. (229.3337,108.0748) .. controls (230.2176,108.0748) and
      (230.7227,108.0518) .. (230.7227,108.0117) -- cycle(230.7070,106.9858) ..
      controls (230.6873,106.9266) and (230.3386,106.9026) .. (229.3127,106.8899) ..
      controls (228.1236,106.8751) and (227.9448,106.8859) .. (227.9448,106.9688) ..
      controls (227.9448,107.0517) and (228.1338,107.0647) .. (229.3390,107.0647) ..
      controls (230.4263,107.0647) and (230.7274,107.0473) .. (230.7070,106.9858) --
      cycle(230.7070,105.9756) .. controls (230.6873,105.9164) and
      (230.3386,105.8924) .. (229.3127,105.8797) .. controls (228.1236,105.8649) and
      (227.9448,105.8757) .. (227.9448,105.9586) .. controls (227.9448,106.0415) and
      (228.1338,106.0545) .. (229.3390,106.0545) .. controls (230.4263,106.0545) and
      (230.7274,106.0371) .. (230.7070,105.9756) -- cycle(230.7070,104.9339) ..
      controls (230.7275,104.8724) and (230.4263,104.8550) .. (229.3390,104.8550) ..
      controls (228.1338,104.8550) and (227.9448,104.8680) .. (227.9448,104.9509) ..
      controls (227.9448,105.0343) and (228.1236,105.0446) .. (229.3127,105.0298) ..
      controls (230.3386,105.0170) and (230.6873,104.9931) .. (230.7070,104.9339) --
      cycle;
  \end{scope}
  \begin{scope}[cm={{0.2703,0.0,0.0,0.2703,(47.69606,6.72892)}}]
    \path[color=black,draw=black,fill=black,line join=miter,line cap=butt,miter
      limit=4.00,even odd rule,line width=0.069pt] (42.8145,21.1191) --
      (42.8145,107.4219) -- (43.4648,107.4219) -- (61.8691,107.4219) --
      (61.8691,21.1191) -- (42.8145,21.1191) -- cycle(44.1152,22.4199) --
      (60.5684,22.4199) -- (60.5684,106.1231) -- (44.1152,106.1231) --
      (44.1152,22.4199) -- cycle;
    \begin{scope}[shift={(-0.20058,0)}]
      \path[draw=black,line join=miter,line cap=butt,miter limit=4.00,even odd
        rule,line width=0.520pt] (46.5085,29.5593) -- (58.5763,29.5593);
      \path[draw=black,line join=miter,line cap=butt,miter limit=4.00,even odd
        rule,line width=0.520pt] (46.5085,33.9720) -- (58.5763,33.9720);
      \path[draw=black,line join=miter,line cap=butt,miter limit=4.00,even odd
        rule,line width=0.520pt] (46.5085,38.3849) -- (58.5763,38.3849);
      \path[draw=black,line join=miter,line cap=butt,miter limit=4.00,even odd
        rule,line width=0.520pt] (46.5085,42.7977) -- (58.5763,42.7977);
    \end{scope}
    \path[fill=black,miter limit=4.00,even odd rule,line width=1.040pt]
      (52.3418,96.7458) circle (0.1588cm);
    \path[cm={{4.33325,0.0,0.0,4.33325,(-941.33439,-425.19092)}},draw=black,fill=cffffff,miter
      limit=4.00,even odd rule,line width=0.015pt] (227.4397,112.9677) --
      (227.4397,103.3397) -- (229.3021,103.3397) -- (231.1646,103.3397) --
      (231.1646,112.9677) -- (231.1646,122.5958) -- (229.3021,122.5958) --
      (227.4397,122.5958) -- (227.4397,112.9677) -- cycle(229.8108,121.6461) ..
      controls (229.8822,121.6197) and (230.0326,121.5206) .. (230.1450,121.4261) ..
      controls (230.9483,120.7501) and (230.6154,119.4131) .. (229.5890,119.1930) ..
      controls (228.6546,118.9926) and (227.8195,119.8427) .. (228.0616,120.7479) ..
      controls (228.2732,121.5391) and (229.0386,121.9321) .. (229.8108,121.6461) --
      cycle(230.7227,108.0117) .. controls (230.7227,107.9715) and
      (230.2176,107.9486) .. (229.3337,107.9486) .. controls (228.4498,107.9486) and
      (227.9447,107.9716) .. (227.9447,108.0117) .. controls (227.9447,108.0519) and
      (228.4498,108.0748) .. (229.3337,108.0748) .. controls (230.2176,108.0748) and
      (230.7227,108.0518) .. (230.7227,108.0117) -- cycle(230.7070,106.9858) ..
      controls (230.6873,106.9266) and (230.3386,106.9026) .. (229.3127,106.8899) ..
      controls (228.1236,106.8751) and (227.9448,106.8859) .. (227.9448,106.9688) ..
      controls (227.9448,107.0517) and (228.1338,107.0647) .. (229.3390,107.0647) ..
      controls (230.4263,107.0647) and (230.7274,107.0473) .. (230.7070,106.9858) --
      cycle(230.7070,105.9756) .. controls (230.6873,105.9164) and
      (230.3386,105.8924) .. (229.3127,105.8797) .. controls (228.1236,105.8649) and
      (227.9448,105.8757) .. (227.9448,105.9586) .. controls (227.9448,106.0415) and
      (228.1338,106.0545) .. (229.3390,106.0545) .. controls (230.4263,106.0545) and
      (230.7274,106.0371) .. (230.7070,105.9756) -- cycle(230.7070,104.9339) ..
      controls (230.7275,104.8724) and (230.4263,104.8550) .. (229.3390,104.8550) ..
      controls (228.1338,104.8550) and (227.9448,104.8680) .. (227.9448,104.9509) ..
      controls (227.9448,105.0343) and (228.1236,105.0446) .. (229.3127,105.0298) ..
      controls (230.3386,105.0170) and (230.6873,104.9931) .. (230.7070,104.9339) --
      cycle;
  \end{scope}
  \path[draw=cffffff,fill=cffffff,line join=miter,line cap=butt,even odd rule,line
    width=0.937pt] (15.4119,10.6233) -- (69.8205,10.6233);
  \path[draw=cffffff,line join=miter,line cap=butt,even odd rule,line
    width=0.937pt] (15.4119,48.2227) -- (69.8205,48.2227);

\end{tikzpicture}

%% file: s2.tex
\definecolor{cffffff}{RGB}{255,255,255}

\begin{tikzpicture}[y=0.80pt, x=0.80pt, yscale=-1.000000, xscale=1.000000, inner sep=0pt, outer sep=0pt]
  \begin{scope}[cm={{0.34694,0.0,0.0,0.34694,(50.72879,8.4793)}}]
    \begin{scope}[shift={(97.33056,-1.65757)}]
      \path[draw=black,line join=miter,line cap=butt,miter limit=4.00,even odd
        rule,line width=0.995pt] (-100.7615,76.7997) -- (-98.2619,79.1724) --
        (-74.7128,79.1724) -- (-71.7028,76.9337) -- (-100.7615,76.7997) -- cycle;
      \path[draw=black,line join=miter,line cap=butt,miter limit=4.00,even odd
        rule,line width=1.040pt] (-95.9120,76.7969) -- (-101.7626,59.0108) --
        (-70.2470,59.0108) -- (-76.8778,76.7189) -- cycle;
      \path[draw=black,line join=miter,line cap=butt,miter limit=4.00,even odd
        rule,line width=1.040pt] (-100.8742,61.0458) .. controls (-100.8742,61.0458)
        and (-103.7304,60.9094) .. (-104.7661,62.3934) .. controls (-105.8018,63.8773)
        and (-103.8855,68.5445) .. (-102.4149,70.1951) .. controls (-100.9443,71.8458)
        and (-98.1728,70.0031) .. (-98.1728,70.0031);
    \end{scope}
    \path[cm={{4.33325,0.0,0.0,4.33325,(-941.33439,-425.19092)}},draw=black,fill=cffffff,miter
      limit=4.00,even odd rule,line width=0.008pt] (216.2849,113.9131) .. controls
      (216.0371,113.8016) and (215.6562,112.9948) .. (215.6202,112.5051) .. controls
      (215.6036,112.2793) and (215.6067,112.2644) .. (215.6876,112.1838) .. controls
      (215.7831,112.0886) and (216.0130,112.0000) .. (216.1646,112.0000) --
      (216.2655,112.0000) -- (216.5640,112.9076) -- (216.8625,113.8152) --
      (216.7473,113.8692) .. controls (216.5785,113.9482) and (216.4004,113.9651) ..
      (216.2849,113.9132) -- cycle;
    \path[cm={{4.33325,0.0,0.0,4.33325,(-941.33439,-425.19092)}},draw=black,fill=cffffff,miter
      limit=4.00,even odd rule,line width=0.008pt] (217.6624,115.2977) .. controls
      (217.6624,115.2887) and (217.3854,114.4392) .. (217.0468,113.4108) .. controls
      (216.7083,112.3823) and (216.4313,111.5305) .. (216.4313,111.5179) .. controls
      (216.4313,111.5053) and (217.9654,111.4950) .. (219.8405,111.4950) .. controls
      (222.0778,111.4950) and (223.2498,111.5058) .. (223.2498,111.5264) .. controls
      (223.2498,111.5437) and (222.9363,112.3937) .. (222.5532,113.4155) --
      (221.8567,115.2732) -- (221.2195,115.2938) .. controls (220.4541,115.3186) and
      (217.6624,115.3216) .. (217.6624,115.2978) -- cycle;
    \path[cm={{4.33325,0.0,0.0,4.33325,(-941.33439,-425.19092)}},draw=black,fill=cffffff,miter
      limit=4.00,even odd rule,line width=0.008pt] (216.9485,115.7486) --
      (216.8065,115.6145) -- (219.6257,115.6235) .. controls (221.1762,115.6285) and
      (222.5016,115.6354) .. (222.5711,115.6393) -- (222.6974,115.6463) --
      (222.5429,115.7647) -- (222.3884,115.8830) -- (219.7394,115.8830) --
      (217.0905,115.8830) -- (216.9485,115.7489) -- cycle;
  \end{scope}
  \begin{scope}[cm={{0.20872,0.0,0.0,0.20872,(-4.15391,74.04193)}}]
  \end{scope}
  \path[xscale=0.998,yscale=1.002,fill=black,line join=miter,line cap=butt,line
    width=0.800pt] (19.5888,46.3275) node[above right] (text7140-9-2-0) {\texttt{B}};
  \path[xscale=0.998,yscale=1.002,fill=black,line join=miter,line cap=butt,line
    width=0.800pt] (58.0972,46.3275) node[above right] (text7140-9-2-0-73) {\texttt{D}};
  \begin{scope}[cm={{0.2703,0.0,0.0,0.2703,(9.40853,6.72892)}}]
    \path[color=black,draw=black,fill=black,line join=miter,line cap=butt,miter
      limit=4.00,even odd rule,line width=0.069pt] (42.8145,21.1191) --
      (42.8145,107.4219) -- (43.4648,107.4219) -- (61.8691,107.4219) --
      (61.8691,21.1191) -- (42.8145,21.1191) -- cycle(44.1152,22.4199) --
      (60.5684,22.4199) -- (60.5684,106.1231) -- (44.1152,106.1231) --
      (44.1152,22.4199) -- cycle;
    \begin{scope}[shift={(-0.20058,0)}]
      \path[draw=black,line join=miter,line cap=butt,miter limit=4.00,even odd
        rule,line width=0.520pt] (46.5085,29.5593) -- (58.5763,29.5593);
      \path[draw=black,line join=miter,line cap=butt,miter limit=4.00,even odd
        rule,line width=0.520pt] (46.5085,33.9720) -- (58.5763,33.9720);
      \path[draw=black,line join=miter,line cap=butt,miter limit=4.00,even odd
        rule,line width=0.520pt] (46.5085,38.3849) -- (58.5763,38.3849);
      \path[draw=black,line join=miter,line cap=butt,miter limit=4.00,even odd
        rule,line width=0.520pt] (46.5085,42.7977) -- (58.5763,42.7977);
    \end{scope}
    \path[fill=black,miter limit=4.00,even odd rule,line width=1.040pt]
      (52.3418,96.7458) circle (0.1588cm);
    \path[cm={{4.33325,0.0,0.0,4.33325,(-941.33439,-425.19092)}},draw=black,fill=cffffff,miter
      limit=4.00,even odd rule,line width=0.015pt] (227.4397,112.9677) --
      (227.4397,103.3397) -- (229.3021,103.3397) -- (231.1646,103.3397) --
      (231.1646,112.9677) -- (231.1646,122.5958) -- (229.3021,122.5958) --
      (227.4397,122.5958) -- (227.4397,112.9677) -- cycle(229.8108,121.6461) ..
      controls (229.8822,121.6197) and (230.0326,121.5206) .. (230.1450,121.4261) ..
      controls (230.9483,120.7501) and (230.6154,119.4131) .. (229.5890,119.1930) ..
      controls (228.6546,118.9926) and (227.8195,119.8427) .. (228.0616,120.7479) ..
      controls (228.2732,121.5391) and (229.0386,121.9321) .. (229.8108,121.6461) --
      cycle(230.7227,108.0117) .. controls (230.7227,107.9715) and
      (230.2176,107.9486) .. (229.3337,107.9486) .. controls (228.4498,107.9486) and
      (227.9447,107.9716) .. (227.9447,108.0117) .. controls (227.9447,108.0519) and
      (228.4498,108.0748) .. (229.3337,108.0748) .. controls (230.2176,108.0748) and
      (230.7227,108.0518) .. (230.7227,108.0117) -- cycle(230.7070,106.9858) ..
      controls (230.6873,106.9266) and (230.3386,106.9026) .. (229.3127,106.8899) ..
      controls (228.1236,106.8751) and (227.9448,106.8859) .. (227.9448,106.9688) ..
      controls (227.9448,107.0517) and (228.1338,107.0647) .. (229.3390,107.0647) ..
      controls (230.4263,107.0647) and (230.7274,107.0473) .. (230.7070,106.9858) --
      cycle(230.7070,105.9756) .. controls (230.6873,105.9164) and
      (230.3386,105.8924) .. (229.3127,105.8797) .. controls (228.1236,105.8649) and
      (227.9448,105.8757) .. (227.9448,105.9586) .. controls (227.9448,106.0415) and
      (228.1338,106.0545) .. (229.3390,106.0545) .. controls (230.4263,106.0545) and
      (230.7274,106.0371) .. (230.7070,105.9756) -- cycle(230.7070,104.9339) ..
      controls (230.7275,104.8724) and (230.4263,104.8550) .. (229.3390,104.8550) ..
      controls (228.1338,104.8550) and (227.9448,104.8680) .. (227.9448,104.9509) ..
      controls (227.9448,105.0343) and (228.1236,105.0446) .. (229.3127,105.0298) ..
      controls (230.3386,105.0170) and (230.6873,104.9931) .. (230.7070,104.9339) --
      cycle;
  \end{scope}
  \begin{scope}[cm={{0.2703,0.0,0.0,0.2703,(47.69606,6.72892)}}]
    \path[color=black,draw=black,fill=black,line join=miter,line cap=butt,miter
      limit=4.00,even odd rule,line width=0.069pt] (42.8145,21.1191) --
      (42.8145,107.4219) -- (43.4648,107.4219) -- (61.8691,107.4219) --
      (61.8691,21.1191) -- (42.8145,21.1191) -- cycle(44.1152,22.4199) --
      (60.5684,22.4199) -- (60.5684,106.1231) -- (44.1152,106.1231) --
      (44.1152,22.4199) -- cycle;
    \begin{scope}[shift={(-0.20058,0)}]
      \path[draw=black,line join=miter,line cap=butt,miter limit=4.00,even odd
        rule,line width=0.520pt] (46.5085,29.5593) -- (58.5763,29.5593);
      \path[draw=black,line join=miter,line cap=butt,miter limit=4.00,even odd
        rule,line width=0.520pt] (46.5085,33.9720) -- (58.5763,33.9720);
      \path[draw=black,line join=miter,line cap=butt,miter limit=4.00,even odd
        rule,line width=0.520pt] (46.5085,38.3849) -- (58.5763,38.3849);
      \path[draw=black,line join=miter,line cap=butt,miter limit=4.00,even odd
        rule,line width=0.520pt] (46.5085,42.7977) -- (58.5763,42.7977);
    \end{scope}
    \path[fill=black,miter limit=4.00,even odd rule,line width=1.040pt]
      (52.3418,96.7458) circle (0.1588cm);
    \path[cm={{4.33325,0.0,0.0,4.33325,(-941.33439,-425.19092)}},draw=black,fill=cffffff,miter
      limit=4.00,even odd rule,line width=0.015pt] (227.4397,112.9677) --
      (227.4397,103.3397) -- (229.3021,103.3397) -- (231.1646,103.3397) --
      (231.1646,112.9677) -- (231.1646,122.5958) -- (229.3021,122.5958) --
      (227.4397,122.5958) -- (227.4397,112.9677) -- cycle(229.8108,121.6461) ..
      controls (229.8822,121.6197) and (230.0326,121.5206) .. (230.1450,121.4261) ..
      controls (230.9483,120.7501) and (230.6154,119.4131) .. (229.5890,119.1930) ..
      controls (228.6546,118.9926) and (227.8195,119.8427) .. (228.0616,120.7479) ..
      controls (228.2732,121.5391) and (229.0386,121.9321) .. (229.8108,121.6461) --
      cycle(230.7227,108.0117) .. controls (230.7227,107.9715) and
      (230.2176,107.9486) .. (229.3337,107.9486) .. controls (228.4498,107.9486) and
      (227.9447,107.9716) .. (227.9447,108.0117) .. controls (227.9447,108.0519) and
      (228.4498,108.0748) .. (229.3337,108.0748) .. controls (230.2176,108.0748) and
      (230.7227,108.0518) .. (230.7227,108.0117) -- cycle(230.7070,106.9858) ..
      controls (230.6873,106.9266) and (230.3386,106.9026) .. (229.3127,106.8899) ..
      controls (228.1236,106.8751) and (227.9448,106.8859) .. (227.9448,106.9688) ..
      controls (227.9448,107.0517) and (228.1338,107.0647) .. (229.3390,107.0647) ..
      controls (230.4263,107.0647) and (230.7274,107.0473) .. (230.7070,106.9858) --
      cycle(230.7070,105.9756) .. controls (230.6873,105.9164) and
      (230.3386,105.8924) .. (229.3127,105.8797) .. controls (228.1236,105.8649) and
      (227.9448,105.8757) .. (227.9448,105.9586) .. controls (227.9448,106.0415) and
      (228.1338,106.0545) .. (229.3390,106.0545) .. controls (230.4263,106.0545) and
      (230.7274,106.0371) .. (230.7070,105.9756) -- cycle(230.7070,104.9339) ..
      controls (230.7275,104.8724) and (230.4263,104.8550) .. (229.3390,104.8550) ..
      controls (228.1338,104.8550) and (227.9448,104.8680) .. (227.9448,104.9509) ..
      controls (227.9448,105.0343) and (228.1236,105.0446) .. (229.3127,105.0298) ..
      controls (230.3386,105.0170) and (230.6873,104.9931) .. (230.7070,104.9339) --
      cycle;
  \end{scope}
  \path[draw=cffffff,fill=cffffff,line join=miter,line cap=butt,even odd rule,line
    width=0.937pt] (15.4119,10.6233) -- (69.8205,10.6233);
  \path[draw=cffffff,line join=miter,line cap=butt,even odd rule,line
    width=0.937pt] (15.4119,48.2227) -- (69.8205,48.2227);
  \path[xscale=0.998,yscale=1.002,fill=black,line join=miter,line cap=butt,line
    width=0.800pt] (50.3730,46.3275) node[above right] (text7140-9-2-0-73-4) {\texttt{C}};

\end{tikzpicture}

%% file: s3.tex
\definecolor{cffffff}{RGB}{255,255,255}

\begin{tikzpicture}[y=0.80pt, x=0.80pt, yscale=-1.000000, xscale=1.000000, inner sep=0pt, outer sep=0pt]
\begin{scope}[cm={{0.20872,0.0,0.0,0.20872,(-4.15391,74.04193)}}]
\end{scope}
  \path[xscale=0.998,yscale=1.002,fill=black,line join=miter,line cap=butt,line
    width=0.800pt] (19.5888,46.3275) node[above right] (text7140-9-2-0) {\texttt{B}};
  \begin{scope}[cm={{0.2703,0.0,0.0,0.2703,(9.40853,6.72892)}}]
    \path[color=black,draw=black,fill=black,line join=miter,line cap=butt,miter
      limit=4.00,even odd rule,line width=0.069pt] (42.8145,21.1191) --
      (42.8145,107.4219) -- (43.4648,107.4219) -- (61.8691,107.4219) --
      (61.8691,21.1191) -- (42.8145,21.1191) -- cycle(44.1152,22.4199) --
      (60.5684,22.4199) -- (60.5684,106.1231) -- (44.1152,106.1231) --
      (44.1152,22.4199) -- cycle;
    \begin{scope}[shift={(-0.20058,0)}]
      \path[draw=black,line join=miter,line cap=butt,miter limit=4.00,even odd
        rule,line width=0.520pt] (46.5085,29.5593) -- (58.5763,29.5593);
      \path[draw=black,line join=miter,line cap=butt,miter limit=4.00,even odd
        rule,line width=0.520pt] (46.5085,33.9720) -- (58.5763,33.9720);
      \path[draw=black,line join=miter,line cap=butt,miter limit=4.00,even odd
        rule,line width=0.520pt] (46.5085,38.3849) -- (58.5763,38.3849);
      \path[draw=black,line join=miter,line cap=butt,miter limit=4.00,even odd
        rule,line width=0.520pt] (46.5085,42.7977) -- (58.5763,42.7977);
    \end{scope}
    \path[fill=black,miter limit=4.00,even odd rule,line width=1.040pt]
      (52.3418,96.7458) circle (0.1588cm);
    \path[cm={{4.33325,0.0,0.0,4.33325,(-941.33439,-425.19092)}},draw=black,fill=cffffff,miter
      limit=4.00,even odd rule,line width=0.015pt] (227.4397,112.9677) --
      (227.4397,103.3397) -- (229.3021,103.3397) -- (231.1646,103.3397) --
      (231.1646,112.9677) -- (231.1646,122.5958) -- (229.3021,122.5958) --
      (227.4397,122.5958) -- (227.4397,112.9677) -- cycle(229.8108,121.6461) ..
      controls (229.8822,121.6197) and (230.0326,121.5206) .. (230.1450,121.4261) ..
      controls (230.9483,120.7501) and (230.6154,119.4131) .. (229.5890,119.1930) ..
      controls (228.6546,118.9926) and (227.8195,119.8427) .. (228.0616,120.7479) ..
      controls (228.2732,121.5391) and (229.0386,121.9321) .. (229.8108,121.6461) --
      cycle(230.7227,108.0117) .. controls (230.7227,107.9715) and
      (230.2176,107.9486) .. (229.3337,107.9486) .. controls (228.4498,107.9486) and
      (227.9447,107.9716) .. (227.9447,108.0117) .. controls (227.9447,108.0519) and
      (228.4498,108.0748) .. (229.3337,108.0748) .. controls (230.2176,108.0748) and
      (230.7227,108.0518) .. (230.7227,108.0117) -- cycle(230.7070,106.9858) ..
      controls (230.6873,106.9266) and (230.3386,106.9026) .. (229.3127,106.8899) ..
      controls (228.1236,106.8751) and (227.9448,106.8859) .. (227.9448,106.9688) ..
      controls (227.9448,107.0517) and (228.1338,107.0647) .. (229.3390,107.0647) ..
      controls (230.4263,107.0647) and (230.7274,107.0473) .. (230.7070,106.9858) --
      cycle(230.7070,105.9756) .. controls (230.6873,105.9164) and
      (230.3386,105.8924) .. (229.3127,105.8797) .. controls (228.1236,105.8649) and
      (227.9448,105.8757) .. (227.9448,105.9586) .. controls (227.9448,106.0415) and
      (228.1338,106.0545) .. (229.3390,106.0545) .. controls (230.4263,106.0545) and
      (230.7274,106.0371) .. (230.7070,105.9756) -- cycle(230.7070,104.9339) ..
      controls (230.7275,104.8724) and (230.4263,104.8550) .. (229.3390,104.8550) ..
      controls (228.1338,104.8550) and (227.9448,104.8680) .. (227.9448,104.9509) ..
      controls (227.9448,105.0343) and (228.1236,105.0446) .. (229.3127,105.0298) ..
      controls (230.3386,105.0170) and (230.6873,104.9931) .. (230.7070,104.9339) --
      cycle;
  \end{scope}
\begin{scope}[shift={(-31.39785,0)}]
  \path[xscale=0.998,yscale=1.002,fill=black,line join=miter,line cap=butt,line
    width=0.800pt] (58.0972,46.3275) node[above right] (text7140-9-2-0-73) {\texttt{D}};
  \begin{scope}[cm={{0.2703,0.0,0.0,0.2703,(47.69606,6.72892)}}]
    \path[color=black,draw=black,fill=black,line join=miter,line cap=butt,miter
      limit=4.00,even odd rule,line width=0.069pt] (42.8145,21.1191) --
      (42.8145,107.4219) -- (43.4648,107.4219) -- (61.8691,107.4219) --
      (61.8691,21.1191) -- (42.8145,21.1191) -- cycle(44.1152,22.4199) --
      (60.5684,22.4199) -- (60.5684,106.1231) -- (44.1152,106.1231) --
      (44.1152,22.4199) -- cycle;
    \begin{scope}[shift={(-0.20058,0)}]
      \path[draw=black,line join=miter,line cap=butt,miter limit=4.00,even odd
        rule,line width=0.520pt] (46.5085,29.5593) -- (58.5763,29.5593);
      \path[draw=black,line join=miter,line cap=butt,miter limit=4.00,even odd
        rule,line width=0.520pt] (46.5085,33.9720) -- (58.5763,33.9720);
      \path[draw=black,line join=miter,line cap=butt,miter limit=4.00,even odd
        rule,line width=0.520pt] (46.5085,38.3849) -- (58.5763,38.3849);
      \path[draw=black,line join=miter,line cap=butt,miter limit=4.00,even odd
        rule,line width=0.520pt] (46.5085,42.7977) -- (58.5763,42.7977);
    \end{scope}
    \path[fill=black,miter limit=4.00,even odd rule,line width=1.040pt]
      (52.3418,96.7458) circle (0.1588cm);
    \path[cm={{4.33325,0.0,0.0,4.33325,(-941.33439,-425.19092)}},draw=black,fill=cffffff,miter
      limit=4.00,even odd rule,line width=0.015pt] (227.4397,112.9677) --
      (227.4397,103.3397) -- (229.3021,103.3397) -- (231.1646,103.3397) --
      (231.1646,112.9677) -- (231.1646,122.5958) -- (229.3021,122.5958) --
      (227.4397,122.5958) -- (227.4397,112.9677) -- cycle(229.8108,121.6461) ..
      controls (229.8822,121.6197) and (230.0326,121.5206) .. (230.1450,121.4261) ..
      controls (230.9483,120.7501) and (230.6154,119.4131) .. (229.5890,119.1930) ..
      controls (228.6546,118.9926) and (227.8195,119.8427) .. (228.0616,120.7479) ..
      controls (228.2732,121.5391) and (229.0386,121.9321) .. (229.8108,121.6461) --
      cycle(230.7227,108.0117) .. controls (230.7227,107.9715) and
      (230.2176,107.9486) .. (229.3337,107.9486) .. controls (228.4498,107.9486) and
      (227.9447,107.9716) .. (227.9447,108.0117) .. controls (227.9447,108.0519) and
      (228.4498,108.0748) .. (229.3337,108.0748) .. controls (230.2176,108.0748) and
      (230.7227,108.0518) .. (230.7227,108.0117) -- cycle(230.7070,106.9858) ..
      controls (230.6873,106.9266) and (230.3386,106.9026) .. (229.3127,106.8899) ..
      controls (228.1236,106.8751) and (227.9448,106.8859) .. (227.9448,106.9688) ..
      controls (227.9448,107.0517) and (228.1338,107.0647) .. (229.3390,107.0647) ..
      controls (230.4263,107.0647) and (230.7274,107.0473) .. (230.7070,106.9858) --
      cycle(230.7070,105.9756) .. controls (230.6873,105.9164) and
      (230.3386,105.8924) .. (229.3127,105.8797) .. controls (228.1236,105.8649) and
      (227.9448,105.8757) .. (227.9448,105.9586) .. controls (227.9448,106.0415) and
      (228.1338,106.0545) .. (229.3390,106.0545) .. controls (230.4263,106.0545) and
      (230.7274,106.0371) .. (230.7070,105.9756) -- cycle(230.7070,104.9339) ..
      controls (230.7275,104.8724) and (230.4263,104.8550) .. (229.3390,104.8550) ..
      controls (228.1338,104.8550) and (227.9448,104.8680) .. (227.9448,104.9509) ..
      controls (227.9448,105.0343) and (228.1236,105.0446) .. (229.3127,105.0298) ..
      controls (230.3386,105.0170) and (230.6873,104.9931) .. (230.7070,104.9339) --
      cycle;
  \end{scope}
\end{scope}
\path[draw=cffffff,fill=cffffff,line join=miter,line cap=butt,even odd rule,line
  width=0.937pt] (15.4119,10.6233) -- (69.8205,10.6233);
\path[draw=cffffff,line join=miter,line cap=butt,even odd rule,line
  width=0.937pt] (15.4119,48.2227) -- (69.8205,48.2227);
  \begin{scope}[cm={{0.34694,0.0,0.0,0.34694,(54.81481,8.04919)}}]
    \begin{scope}[shift={(97.33056,-1.65757)}]
      \path[draw=black,line join=miter,line cap=butt,miter limit=4.00,even odd
        rule,line width=0.995pt] (-100.7615,76.7997) -- (-98.2619,79.1724) --
        (-74.7128,79.1724) -- (-71.7028,76.9337) -- (-100.7615,76.7997) -- cycle;
      \path[draw=black,line join=miter,line cap=butt,miter limit=4.00,even odd
        rule,line width=1.040pt] (-95.9120,76.7969) -- (-101.7626,59.0108) --
        (-70.2470,59.0108) -- (-76.8778,76.7189) -- cycle;
      \path[draw=black,line join=miter,line cap=butt,miter limit=4.00,even odd
        rule,line width=1.040pt] (-100.8742,61.0458) .. controls (-100.8742,61.0458)
        and (-103.7304,60.9094) .. (-104.7661,62.3934) .. controls (-105.8018,63.8773)
        and (-103.8855,68.5445) .. (-102.4149,70.1951) .. controls (-100.9443,71.8458)
        and (-98.1728,70.0031) .. (-98.1728,70.0031);
    \end{scope}
    \path[cm={{4.33325,0.0,0.0,4.33325,(-941.33439,-425.19092)}},draw=black,fill=cffffff,miter
      limit=4.00,even odd rule,line width=0.008pt] (216.2849,113.9131) .. controls
      (216.0371,113.8016) and (215.6562,112.9948) .. (215.6202,112.5051) .. controls
      (215.6036,112.2793) and (215.6067,112.2644) .. (215.6876,112.1838) .. controls
      (215.7831,112.0886) and (216.0130,112.0000) .. (216.1646,112.0000) --
      (216.2655,112.0000) -- (216.5640,112.9076) -- (216.8625,113.8152) --
      (216.7473,113.8692) .. controls (216.5785,113.9482) and (216.4004,113.9651) ..
      (216.2849,113.9132) -- cycle;
    \path[cm={{4.33325,0.0,0.0,4.33325,(-941.33439,-425.19092)}},draw=black,fill=cffffff,miter
      limit=4.00,even odd rule,line width=0.008pt] (217.6624,115.2977) .. controls
      (217.6624,115.2887) and (217.3854,114.4392) .. (217.0468,113.4108) .. controls
      (216.7083,112.3823) and (216.4313,111.5305) .. (216.4313,111.5179) .. controls
      (216.4313,111.5053) and (217.9654,111.4950) .. (219.8405,111.4950) .. controls
      (222.0778,111.4950) and (223.2498,111.5058) .. (223.2498,111.5264) .. controls
      (223.2498,111.5437) and (222.9363,112.3937) .. (222.5532,113.4155) --
      (221.8567,115.2732) -- (221.2195,115.2938) .. controls (220.4541,115.3186) and
      (217.6624,115.3216) .. (217.6624,115.2978) -- cycle;
    \path[cm={{4.33325,0.0,0.0,4.33325,(-941.33439,-425.19092)}},draw=black,fill=cffffff,miter
      limit=4.00,even odd rule,line width=0.008pt] (216.9485,115.7486) --
      (216.8065,115.6145) -- (219.6257,115.6235) .. controls (221.1762,115.6285) and
      (222.5016,115.6354) .. (222.5711,115.6393) -- (222.6974,115.6463) --
      (222.5429,115.7647) -- (222.3884,115.8830) -- (219.7394,115.8830) --
      (217.0905,115.8830) -- (216.9485,115.7489) -- cycle;
  \end{scope}
  \path[xscale=0.998,yscale=1.002,fill=black,line join=miter,line cap=butt,line
    width=0.800pt] (54.4672,45.8983) node[above right] (text7140-9-2-0-73-4) {\texttt{C}};

\end{tikzpicture}

%% file: hierarchy.tex
% Graphic for TeX using PGF
% Title: /home/bubx/pubblication/[17 AIRO-AIxIA] fuzzySIT/image/imageBuilder/hierarchy.dia
% Creator: Dia v0.97.3
% CreationDate: Wed Sep 13 13:41:57 2017
% For: bubx
% \usepackage{tikz}
% The following commands are not supported in PSTricks at present
% We define them conditionally, so when they are implemented,
% this pgf file will use them.
\ifx\du\undefined
  \newlength{\du}
\fi
\setlength{\du}{15\unitlength}
\begin{tikzpicture}
\pgftransformxscale{1.000000}
\pgftransformyscale{-1.000000}
\definecolor{dialinecolor}{rgb}{0.000000, 0.000000, 0.000000}
\pgfsetstrokecolor{dialinecolor}
\definecolor{dialinecolor}{rgb}{1.000000, 1.000000, 1.000000}
\pgfsetfillcolor{dialinecolor}
\pgfsetlinewidth{0.100000\du}
\pgfsetdash{}{0pt}
\pgfsetdash{}{0pt}
\pgfsetbuttcap
{
\definecolor{dialinecolor}{rgb}{0.000000, 0.000000, 0.000000}
\pgfsetfillcolor{dialinecolor}
% was here!!!
\pgfsetarrowsend{latex}
\definecolor{dialinecolor}{rgb}{0.000000, 0.000000, 0.000000}
\pgfsetstrokecolor{dialinecolor}
\draw (16.970635\du,9.318135\du)--(19.036923\du,8.377565\du);
}
\pgfsetlinewidth{0.100000\du}
\pgfsetdash{}{0pt}
\pgfsetdash{}{0pt}
\pgfsetbuttcap
{
\definecolor{dialinecolor}{rgb}{0.000000, 0.000000, 0.000000}
\pgfsetfillcolor{dialinecolor}
% was here!!!
\pgfsetarrowsend{latex}
\definecolor{dialinecolor}{rgb}{0.000000, 0.000000, 0.000000}
\pgfsetstrokecolor{dialinecolor}
\draw (23.345630\du,9.318135\du)--(21.279342\du,8.377565\du);
}
\pgfsetlinewidth{0.100000\du}
\pgfsetdash{}{0pt}
\pgfsetdash{}{0pt}
\pgfsetbuttcap
{
\definecolor{dialinecolor}{rgb}{0.000000, 0.000000, 0.000000}
\pgfsetfillcolor{dialinecolor}
% was here!!!
\pgfsetarrowsend{latex}
\definecolor{dialinecolor}{rgb}{0.000000, 0.000000, 0.000000}
\pgfsetstrokecolor{dialinecolor}
\draw (18.435566\du,9.766102\du)--(21.880700\du,9.766102\du);
}
\pgfsetlinewidth{0.100000\du}
\pgfsetdash{}{0pt}
\pgfsetdash{}{0pt}
\pgfsetbuttcap
{
\definecolor{dialinecolor}{rgb}{0.000000, 0.000000, 0.000000}
\pgfsetfillcolor{dialinecolor}
% was here!!!
\pgfsetarrowsend{latex}
\definecolor{dialinecolor}{rgb}{0.000000, 0.000000, 0.000000}
\pgfsetstrokecolor{dialinecolor}
\draw (21.880700\du,10.321505\du)--(18.435566\du,10.321505\du);
}
% setfont left to latex
\definecolor{dialinecolor}{rgb}{0.000000, 0.000000, 0.000000}
\pgfsetstrokecolor{dialinecolor}
\node[anchor=east] at (18.003779\du,8.369294\du){1.0};
% setfont left to latex
\definecolor{dialinecolor}{rgb}{0.000000, 0.000000, 0.000000}
\pgfsetstrokecolor{dialinecolor}
\node[anchor=east] at (18.003779\du,8.757350\du){};
% setfont left to latex
\definecolor{dialinecolor}{rgb}{0.000000, 0.000000, 0.000000}
\pgfsetstrokecolor{dialinecolor}
\node[anchor=west] at (22.312486\du,8.369294\du){1.0};
% setfont left to latex
\definecolor{dialinecolor}{rgb}{0.000000, 0.000000, 0.000000}
\pgfsetstrokecolor{dialinecolor}
\node[anchor=west] at (22.312486\du,8.757350\du){};
% setfont left to latex
\definecolor{dialinecolor}{rgb}{0.000000, 0.000000, 0.000000}
\pgfsetstrokecolor{dialinecolor}
\node at (20.158133\du,9.305546\du){0.32};
% setfont left to latex
\definecolor{dialinecolor}{rgb}{0.000000, 0.000000, 0.000000}
\pgfsetstrokecolor{dialinecolor}
\node at (20.158133\du,9.693602\du){};
% setfont left to latex
\definecolor{dialinecolor}{rgb}{0.000000, 0.000000, 0.000000}
\pgfsetstrokecolor{dialinecolor}
\node at (20.158133\du,10.618005\du){};
% setfont left to latex
\definecolor{dialinecolor}{rgb}{0.000000, 0.000000, 0.000000}
\pgfsetstrokecolor{dialinecolor}
\node at (20.158133\du,11.006060\du){0.24};
\definecolor{dialinecolor}{rgb}{1.000000, 1.000000, 1.000000}
\pgfsetfillcolor{dialinecolor}
\pgfpathellipse{\pgfpoint{20.158133\du}{7.864439\du}}{\pgfpoint{1.585630\du}{0\du}}{\pgfpoint{0\du}{0.725669\du}}
\pgfusepath{fill}
\pgfsetlinewidth{0.100000\du}
\pgfsetdash{}{0pt}
\pgfsetdash{}{0pt}
\pgfsetmiterjoin
\definecolor{dialinecolor}{rgb}{0.000000, 0.000000, 0.000000}
\pgfsetstrokecolor{dialinecolor}
\pgfpathellipse{\pgfpoint{20.158133\du}{7.864439\du}}{\pgfpoint{1.585630\du}{0\du}}{\pgfpoint{0\du}{0.725669\du}}
\pgfusepath{stroke}
% setfont left to latex
\definecolor{dialinecolor}{rgb}{0.000000, 0.000000, 0.000000}
\pgfsetstrokecolor{dialinecolor}
\node at (20.158133\du,7.957911\du){$\texttt{Scene}_{\texttt{1}}$};
\definecolor{dialinecolor}{rgb}{1.000000, 1.000000, 1.000000}
\pgfsetfillcolor{dialinecolor}
\pgfpathellipse{\pgfpoint{23.345630\du}{10.043803\du}}{\pgfpoint{1.585630\du}{0\du}}{\pgfpoint{0\du}{0.725669\du}}
\pgfusepath{fill}
\pgfsetlinewidth{0.100000\du}
\pgfsetdash{}{0pt}
\pgfsetdash{}{0pt}
\pgfsetmiterjoin
\definecolor{dialinecolor}{rgb}{0.000000, 0.000000, 0.000000}
\pgfsetstrokecolor{dialinecolor}
\pgfpathellipse{\pgfpoint{23.345630\du}{10.043803\du}}{\pgfpoint{1.585630\du}{0\du}}{\pgfpoint{0\du}{0.725669\du}}
\pgfusepath{stroke}
% setfont left to latex
\definecolor{dialinecolor}{rgb}{0.000000, 0.000000, 0.000000}
\pgfsetstrokecolor{dialinecolor}
\node at (23.345630\du,10.137276\du){$\texttt{Scene}_{\texttt{3}}$};
\definecolor{dialinecolor}{rgb}{1.000000, 1.000000, 1.000000}
\pgfsetfillcolor{dialinecolor}
\pgfpathellipse{\pgfpoint{16.970635\du}{10.043803\du}}{\pgfpoint{1.585630\du}{0\du}}{\pgfpoint{0\du}{0.725669\du}}
\pgfusepath{fill}
\pgfsetlinewidth{0.100000\du}
\pgfsetdash{}{0pt}
\pgfsetdash{}{0pt}
\pgfsetmiterjoin
\definecolor{dialinecolor}{rgb}{0.000000, 0.000000, 0.000000}
\pgfsetstrokecolor{dialinecolor}
\pgfpathellipse{\pgfpoint{16.970635\du}{10.043803\du}}{\pgfpoint{1.585630\du}{0\du}}{\pgfpoint{0\du}{0.725669\du}}
\pgfusepath{stroke}
% setfont left to latex
\definecolor{dialinecolor}{rgb}{0.000000, 0.000000, 0.000000}
\pgfsetstrokecolor{dialinecolor}
\node at (16.970635\du,10.137276\du){$\texttt{Scene}_{\texttt{2}}$};
\end{tikzpicture}